\documentclass[sigconf,nonacm]{acmart}
\AtBeginDocument{}
\acmConference[SIGSPATIAL '26]{The 34th ACM International Conference on
  Advances in Geographic Information Systems}{November 2026}{TBD}

\begin{document}

\title{An Autonomous GeoAI Agent for Arctic Eco-Navigation}

\author{Samira Alkaee Taleghan}
\affiliation{%
  \institution{University of Colorado Denver}
  \city{Denver}
  \state{Colorado}
  \country{USA}}
\email{samira.alkaeetaleghan@ucdenver.edu}

\author{Younghyun Koo}
\affiliation{%
  \institution{National Snow and Ice Data Center (NSIDC), CIRES, University of Colorado Boulder}
  \city{Boulder}
  \state{Colorado}
  \country{USA}}
\email{younghyun.koo@colorado.edu}

\author{Farnoush Banaei-Kashani}
\affiliation{%
  \institution{University of Colorado Denver}
  \city{Denver}
  \state{Colorado}
  \country{USA}}
\email{farnoush.banaei-kashani@ucdenver.edu}
\renewcommand{\shortauthors}{Alkaee Taleghan et al.}

\begin{abstract}
Arctic maritime navigation is becoming increasingly important as
changing sea-ice conditions expand seasonal accessibility while
simultaneously introducing substantial operational, environmental,
and community risks. Arctic route planning is inherently a
multi-criteria problem: routes that improve vessel safety or efficiency
may increase exposure to sea ice, sensitive ecosystems, or nearby
communities. Existing routing methods prioritize travel time, fuel use, and navigational risk, often overlooking ecological and community impacts. We introduce a
human-in-the-loop, multi-agent GeoAI system for Arctic
eco-navigation that integrates operational, physical, ecological, and
community-related criteria within a unified routing framework.
Multiple specialized agents coordinate geospatial data acquisition
and preparation, multi-objective route generation, and skyline-based
decision support. The ecological criteria explicitly account for exposure to sensitive areas, including
Essential Fish Habitat and seal critical habitat. By considering these
ecosystem impacts and potential community burdens while keeping
consequential value judgments under human control, the framework
supports safer, more transparent, and socially responsible Arctic
navigation.
Project page and code are publicly available.\footnote{\url{https://samiraat.github.io/Arctic-Eco-Navigation-Agent/}}
\footnote{\url{https://github.com/samiraat/Arctic-Eco-Navigation-Agent}}
\vspace{-3mm}
\end{abstract}
\vspace{-3mm}


\maketitle


%
%

\begin{figure*}[t]
    \centering
    \includegraphics[width=0.7\linewidth]{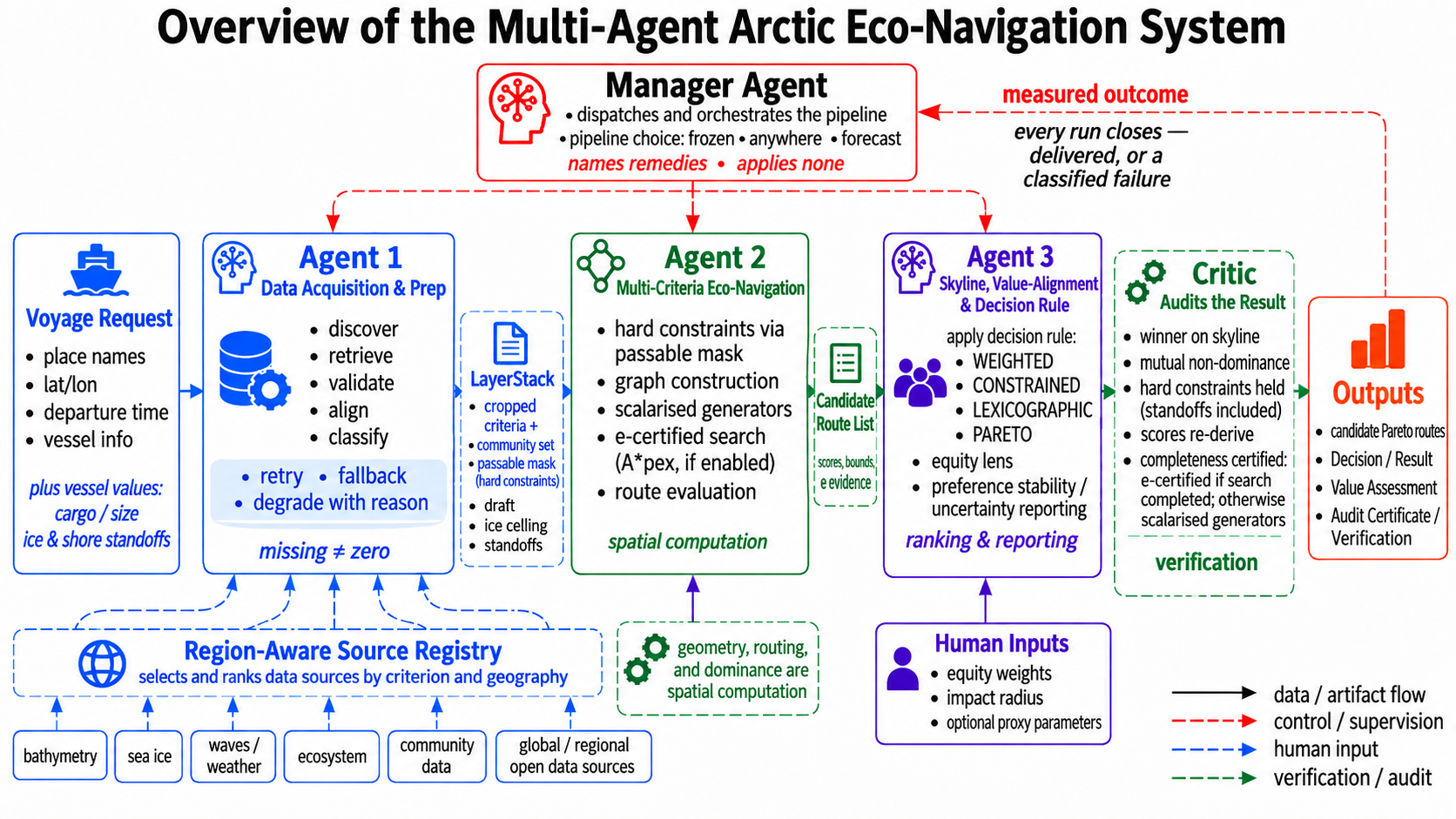}
    \caption{System Overview.}
    
    \label{fig:overview}
\end{figure*}
\section{Problem Statement}
Arctic route planning requires balancing vessel safety and efficiency with
environmental protection and impacts on nearby communities. These objectives
can conflict: shorter routes may increase exposure to sea ice, shallow water,
sensitive habitat, or communities. The problem is therefore multi-criteria,
with no universally ``best'' route. Our goal is to support either one
preference-defined optimum or a set of defensible trade-offs, while keeping
preference and equity judgments under human control.

Formally, a voyage request is $q=(o,d,t,v)$, where $o$ and $d$ are the origin
and destination, $t$ the departure time, and $v$ the vessel characteristics
relevant to navigation, including draft and safety margin. Navigable water is
represented as a graph $G=(V,E)$. Feasibility is enforced before optimization
because some hazards are not treated as trade-offs: fuel or hazardous cargo
released under sea ice can defeat mechanical spill recovery, while large
vessels maintain shoreline standoffs to reduce disturbance to marine mammals
and nearby coastal communities. Each feasible route $r$ is evaluated by
$\mathbf{c}(r)=\bigl(c_1(r),\ldots,c_k(r)\bigr)$ over six criteria:
(1) distance; (2) shallow-water risk from reduced under-keel clearance;
(3) sea-ice difficulty; (4) sea-state exposure to adverse wave and wind
conditions; (5) and ecological exposure to critical habitat and seasonal
Essential Fish Habitat.

The requested output determines the optimization problem. If the user wants
one route, a single criterion is minimized directly, or multiple criteria are
normalized and combined using default or user-assigned weights into one
scalar cost. If the user instead wants the trade-offs, the criteria remain
separate and routing seeks non-dominated routes. The user then selects exact
Pareto search or an $\varepsilon$-approximate search; the latter exposes the
accuracy--computation trade-off through an explicit approximation parameter
and time budget. Thus the choice of one route versus a Pareto set, criterion
selection, weights, and approximation tolerance remain human inputs rather
than values inferred by the system.

The formulation is not tied to a region: the agent accepts endpoints as names
or coordinates, selects the analysis CRS per corridor (EPSG:3413 above
60$^\circ$N, otherwise an azimuthal equidistant projection centred on the
corridor), and re-decides the active criterion set from available evidence.
For our experiments we use the Bering Strait region: a coastal supply vessel
(draft 6\,m plus a 2\,m safety margin, 12\,kn) sailing Nome$\to$Kotzebue,
Alaska, on a 3\,km grid within a 250\,km corridor buffer. Existing ice-routing
systems primarily optimize vessel time, fuel, and risk; our formulation adds
physical, ecological, and community criteria and lets the user request either
a preference-defined optimum or their non-dominated trade-offs.
\subsection{Definitions}

\paragraph{Optimality.}
Because the routing criteria are incommensurable, we define optimality as
Pareto non-dominance rather than a single scalar minimum. A feasible route
$r$ is non-dominated if there is no feasible $r'$ with
$c_i(r') \le c_i(r)$ for every active criterion $i$ and
$c_i(r') < c_i(r)$ for at least one. The system retains this skyline rather
than asserting a single universally ``best'' route. A preferred route may then be identified using stakeholder-supplied weights,
constraints, or priority orderings, or the non-dominated routes may be
presented directly for human comparison.
Thus, any selection beyond Pareto non-dominance reflects stakeholder
preferences rather than an autonomous system judgment.

\paragraph{Vulnerability.}
We define vulnerability as the sensitivity of ecosystems or communities to
potential impacts from vessel traffic. It is distinct from exposure, which
measures how much a route interacts with sensitive areas or nearby
communities. For example, vulnerability may reflect seasonal Essential Fish
Habitat, seal critical habitat, or areas important for subsistence hunting.
These sensitivities can influence ecological and community criteria or, when
a risk is considered unacceptable, motivate hard constraints such as
standoff distances from sea ice or shorelines.

\paragraph{Equity.}
Equity concerns the distribution of route-related burden across communities
rather than only its total. Let $b_j(r)$ denote the burden that route $r$
imposes on community $j$. Stakeholders may assess this distribution using
different principles, including minimizing total burden, minimizing the
maximum burden experienced by any community, or giving greater priority to
socially sensitive communities. These alternatives represent different
conceptions of fairness and may produce different preferred outcomes.
Because selecting among them is a normative judgment, the equity assessment
is kept separate from the multi-criteria route ranking and its definition is
left to stakeholders.

\begin{figure}[htbp]
    \centering
    \includegraphics[width=0.80\linewidth]{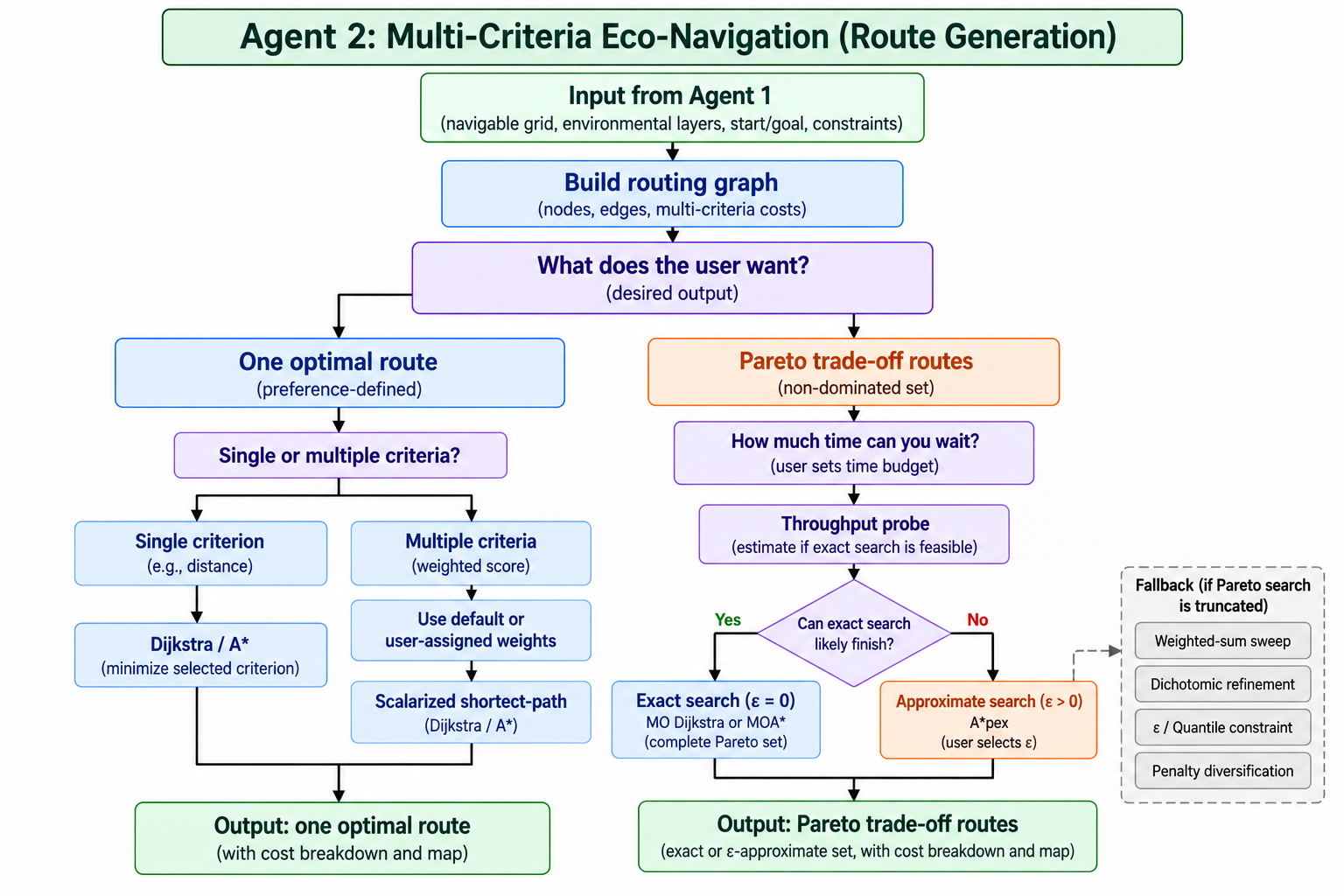}
    \caption{Agent~2 search selection}
    \label{fig:agent2}
\end{figure}

\begin{figure}[htbp]
    \centering
    \includegraphics[width=0.80\linewidth]{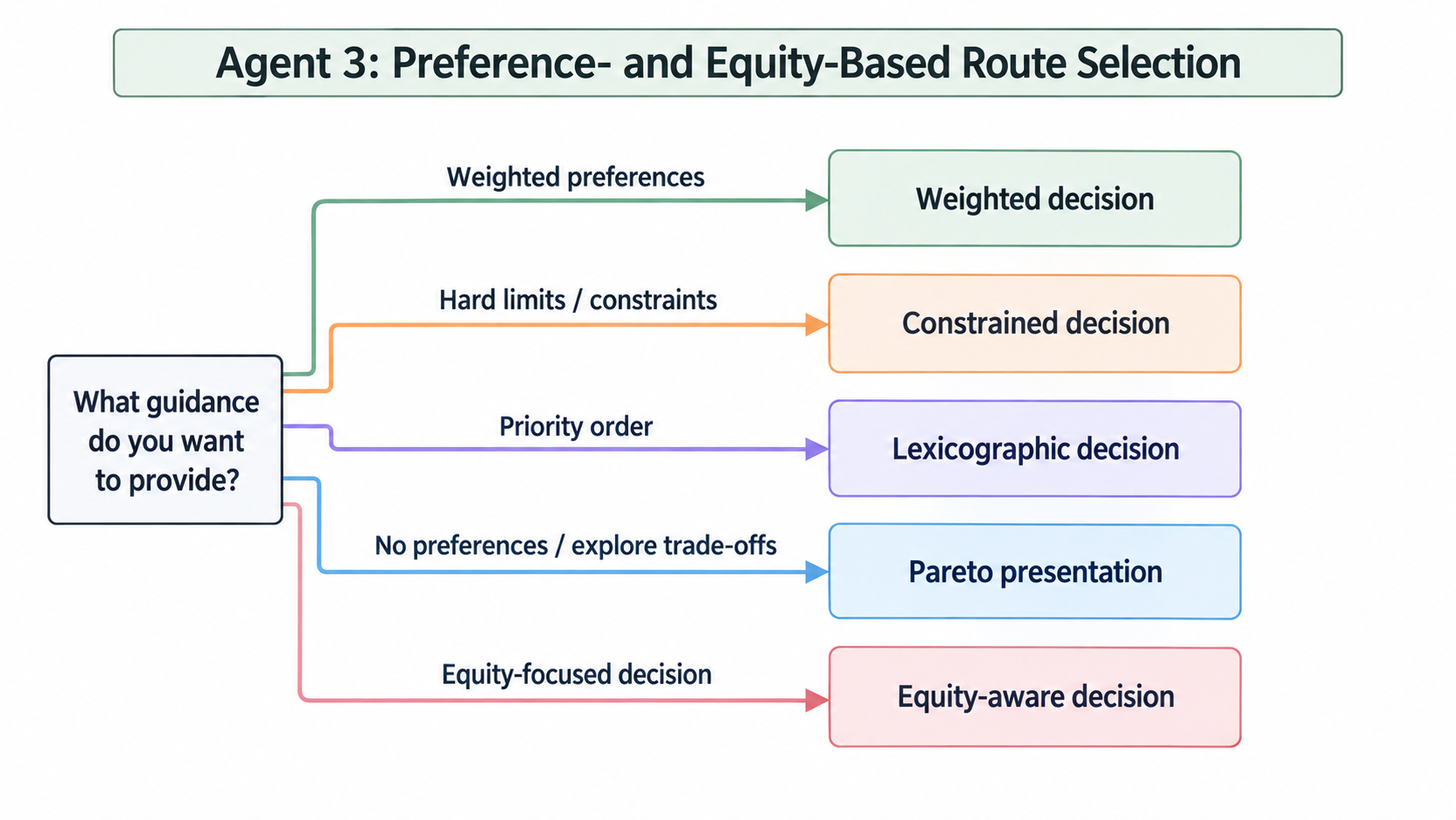}
    \caption{Agent~3 decision rules}
    \label{fig:agent3}
    \vspace{-3mm}
\end{figure}
\vspace{-3mm}
\section{Agent Design}
\label{sec:agent-design}

\paragraph{Architecture and Orchestration}
The system is organised as a set of specialised agents that exchange
information through predefined structured interfaces
(Fig.~\ref{fig:overview}).An orchestration layer first interprets the voyage request and elicits the routing objective. The user chooses either a single optimal route or Pareto trade-offs. Single-route requests use one criterion or human-approved multi-criteria weights, while Pareto requests use exact or approximate search with a user-specified tolerance $\varepsilon$. A computation-time budget is used to assess whether exact search is practical.
Agent~1 acquires and validates geospatial data; Agent~2 constructs the graph and performs the selected routing search; and Agent~3 evaluates the resulting route or Pareto trade-offs and conducts a separate equity analysis. An independent critic verifies the result before release.
The orchestration layer is thin: deterministic checks select eligible processing paths, while the language model interprets requests and elicits missing human choices. Closed agent interfaces prevent it from assigning normative parameters without human approval. The system uses the local open-weights model \texttt{llama3.1:8b} served through Ollama.

\paragraph{Data Acquisition and Preparation}
Agent~1 integrates open geospatial data from USNIC for archived sea ice,
the NOAA Global Forecast System (GFS) for forecast sea ice, NOAA NCEI for
bathymetry, NOAA Fisheries for Essential Fish Habitat and seal critical
habitat, Open-Meteo for waves and wind, and Alaska DCRA and BOEM for
community and subsistence information. Candidate sources are discovered
through catalogue search restricted to an allowlist of trusted providers,
then retrieved, validated, and aligned to a common metre-based projected
grid---EPSG:3413 for Arctic corridors and a corridor-centred azimuthal
equidistant projection otherwise. Because clipping is performed in the
projected frame, the antimeridian requires no special case. Coordinate
reference systems are read from each source and never assumed; an undeclared
CRS raises an error rather than defaulting to EPSG:4326. Polygon-to-grid alignment supports both cell-centre rasterisation and
fractional cell coverage estimated by supersampling; the latter is kept
as an explicit analysis option rather than silently replacing the
baseline spatial representation. Coverage is evaluated
over navigable water rather than the full grid; criteria below a 50\%
evidence floor are demoted to low confidence, while those with no usable
coverage are dropped and reported rather than zero-filled. At the cell level,
absence is classified as measured, ice-attributed, or unexplained; only
unexplained cells receive an upper uncertainty bound based on the worst
observed value on navigable water, so missing data cannot create a dominance
claim. Acquisition failures follow a predefined sequence of retry,
age-stamped stored copy with reduced confidence, degradation with a caveat,
and criterion removal; loss of a required criterion stops the run. Forecast
layers must cover the requested departure time, otherwise the system returns
\textsc{not\_covered} rather than substituting archived conditions.

\paragraph{Routing Optimisation}
Agent~2 separates single-solution optimisation from Pareto search. It builds a 16-connected graph over navigable cells, removing infeasible nodes or edges using hard constraints such as land, depth, connectivity, ice limits, and standoff rules. Ecological and cultural exposure remain optimisation criteria rather than hard constraints.
For a single criterion, Agent~2 solves a standard shortest-path problem
with that criterion as the edge cost. For multiple criteria with one
preferred route, the active criteria are normalised via fixed anchors and
combined additively with human-approved weights,
\(J(r)=\sum_i w_i \hat{C}_i(r)\).
after which a conventional shortest-path algorithm returns the route that is
optimal for that stated scalar objective. The interface presents predefined
weights as defaults but requires the human either to accept them or provide
alternative values; weights are never inferred silently by the language
model.
If the user instead requests the trade-off set, the criteria remain separate
and Agent~2 performs Pareto multi-objective search. Exact search uses
$\varepsilon=0$ and targets the complete Pareto set; the implementation
supports multi-objective Dijkstra~\cite{martins1984multicriteria} and
multi-objective A*~\cite{mandow2010namoa}, the latter using per-criterion
lower bounds to reduce unnecessary expansions. Approximate search uses
A*pex~\cite{zhang2022apex} with a human-approved $\varepsilon>0$ and, when
the search completes, returns a certified $\varepsilon$-approximate Pareto
set.
The user provides a computation-time budget, and a throughput probe estimates whether exact Pareto search is practical. If the search is truncated, no completeness or approximation guarantee is claimed; fallback generators may augment the partial result using weighted-sum, Aneja--Nair refinement~\cite{aneja1979bicriteria}, quantile constraints, and edge-penalty diversification (Fig.~\ref{fig:agent2}). Exact bi-objective BOA*~\cite{ulloa2020boa} is used for offline validation.

\paragraph{Route Presentation and Equity Analysis}
Agent~3 adapts to the routing request rather than applying a single ranking procedure to every run. When one optimal route is requested, Agent~2 has already optimised the stated objective, so Agent~3 simply reports that route with its raw per-criterion values and provenance, without re-weighting it.
When the user requests Pareto trade-offs, Agent~3 receives the exact or
$\varepsilon$-approximate Pareto set produced by Agent~2 and presents the
non-dominated alternatives without inventing a preferred route. Agent~3 first removes spatially duplicate routes using the configured geometric similarity criterion, then applies dominance filtering to the remaining routes. If a
multi-objective search is truncated and fallback candidate generation is
used, Agent~3 computes the candidate skyline using pairwise dominance
filtering~\cite{borzsonyi2001skyline}; this skyline is explicitly described
as exact only with respect to the generated candidate pool and not as the
complete Pareto frontier.
The equity lens is applied independently of the technical optimisation. It
evaluates community burdens for the returned route or route set under
human-selected equity assumptions and reports \textsc{agreement},
\textsc{conflict}, \textsc{tie}, or \textsc{undetermined}. Technical and
equity values are never collapsed into a single composite score.

\paragraph{Independent Critic and Verification}
Before release, a deterministic critic verifies route feasibility and branch-specific results. It checks the reported optimum for single-route searches and non-dominance, $\varepsilon$, and search guarantees for Pareto searches. It also verifies provenance, standoff constraints, and equity invariants where applicable. Any failed or crashed check causes verification failure, and no language model is used in this stage.

\section{Results}\label{sec:results}

\begin{figure}[t]
  \centering
  \includegraphics[width=\linewidth]{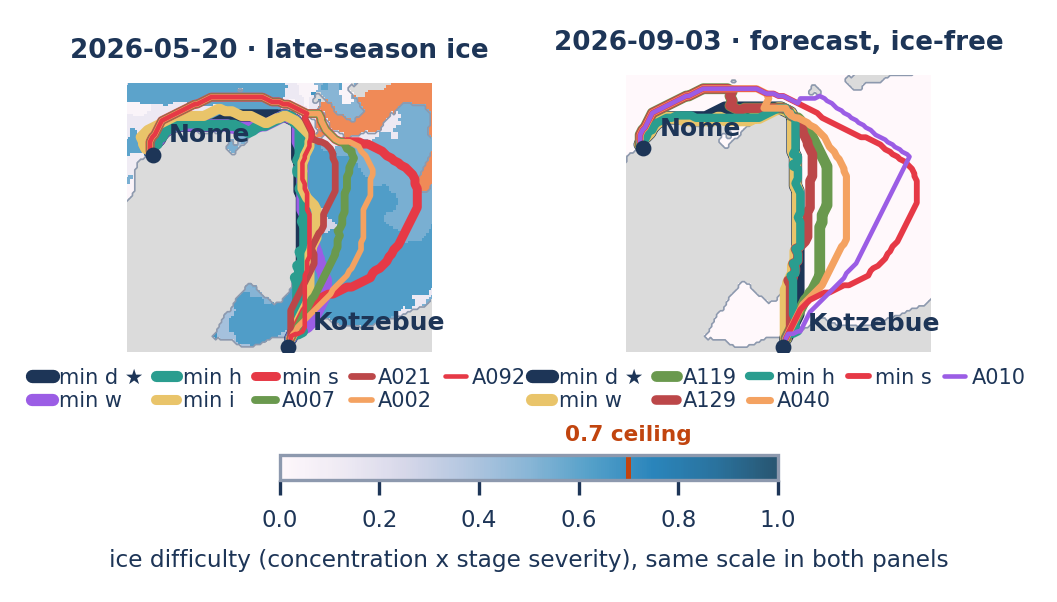}
  \caption{Two departures on identical settings, both panels shaded by ice
    difficulty on one fixed 0--1 scale: 2026-05-20 from the USNIC chart
    (left), 2026-09-03 from the GFS forecast (right).}
  \label{fig:map}
\end{figure}

\begin{figure}[t]
  \centering
  \includegraphics[width=\linewidth]{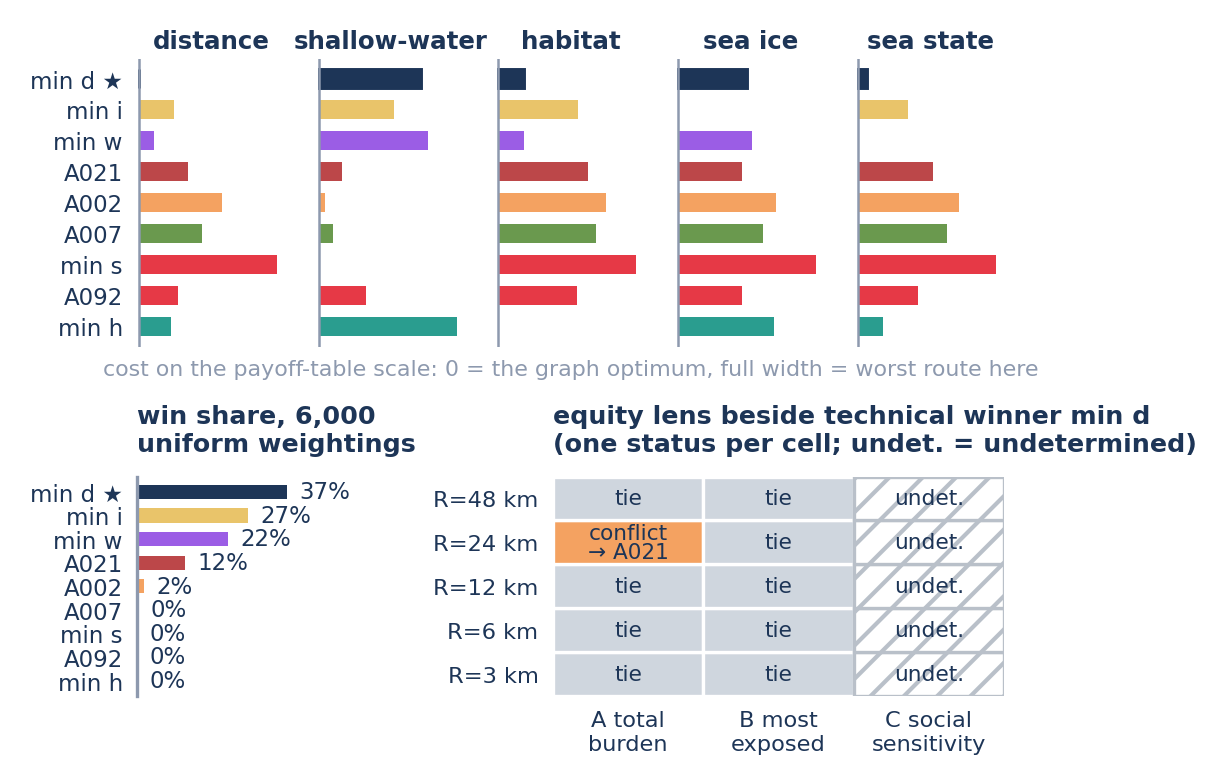}
  \caption{Preference and equity sensitivity for the 2026-05-20 run.}
  \label{fig:sens}
  \vspace{-7mm}
\end{figure}

\begin{table}[t]
  \caption{$\varepsilon$-approximate Pareto routes at $\varepsilon=0.35$.}
  \label{tab:skyline}
  \footnotesize
  \begin{tabular}{lrrrrr}
    \toprule
    Route & dist. & shallow & habitat & ice & sea state \\
          & nmi & risk-nmi & cat.-nmi & diff.-nmi & nmi \\
    \midrule
    \multicolumn{6}{@{}l}{\emph{2026-05-20, ice season }} \\
    A000 & 377.7 & \textbf{93.0} & 487.2 & 177.0 & 66.4 \\
    A002 & 332.7 & 98.1 & 442.8 & 153.6 & 59.6 \\
    A007 & 316.6 & 104.3 & 426.8 & 145.9 & 57.3 \\
    A021$\star$ & 304.9 & 112.2 & 415.7 & \textbf{134.1} & 54.8 \\
    A092 & \textbf{296.7} & 131.5 & \textbf{399.3} & 134.2 & \textbf{52.2} \\
    \midrule
    \multicolumn{6}{@{}l}{\emph{2026-09-03, forecast (ice-free)}} \\
    A000 & 372.2 & \textbf{92.5} & 480.0 & 0.0 & 174.6 \\
    A010 & 367.9 & 102.1 & 408.1 & 0.0 & 173.1 \\
    A040 & 314.8 & 112.3 & 413.5 & 0.0 & 147.0 \\
    A119$\star$ & 301.3 & 127.0 & 385.8 & 0.0 & 138.1 \\
    A129 & \textbf{300.6} & 152.8 & \textbf{374.3} & 0.0 & \textbf{133.9} \\
    \bottomrule
  \end{tabular}
\end{table}

\begin{table}[t]
  \caption{A*pex runtime and certification across $\varepsilon$ values.}
  \label{tab:certfrontier}
  \footnotesize
  \begin{tabular}{lrrrl}
    \toprule
    $\varepsilon$ & seconds & expansions & reps. & status \\
    \midrule
    0.10 & 120 & 807{,}500 & 55 & incomplete \\
    0.25 & 5{,}401 & 6{,}308{,}500 & 467 & incomplete \\
    0.35 & 371 & 3{,}171{,}342 & 229 & complete \\
    0.50 & 135 & 1{,}968{,}725 & 5 & complete \\
    1.00 & 0.3 & 148 & 1 & complete \\
    \midrule
    \multicolumn{5}{@{}l}{\emph{2026-09-03}} \\
    0.35 & 715 & 2{,}208{,}721 & 176 & complete \\
    0.50 & 29.6 & 375{,}953 & 7 & complete \\
    \bottomrule
  \end{tabular}
\end{table}
Two departures on identical settings define the instances
(Fig.~\ref{fig:map}): 2026-05-20, when ice covered 89\% of navigable water
and five criteria rank, and 2026-09-03, when ice is measured zero and dropped,
leaving four criteria. On both, A*pex at $\varepsilon=0.35$ completes within
minutes (371\,s and 715\,s), so every feasible route is within 35\% of a
returned route on every ranking criterion (Table~\ref{tab:skyline}).
An $\varepsilon$-Pareto set provides coverage but may omit extreme solutions,
so each set is augmented with its single-criterion optima. This can only widen
coverage and therefore preserves the approximation guarantee. After spatial
deduplication at 9\,km, the augmented sets contain nine and eight distinct
corridors (Table~\ref{tab:certfrontier}), with no pair sharing more than 59\%
of their cells.
The results are internally consistent: rebuilt
graphs reproduce every recorded criterion total to $\leq
6.5\times10^{-9}$ relative, all representatives are mutually
non-dominated, and the certificate covers the system's earlier output with
margin (measured $\varepsilon \leq 0.282$). Scored on the ranking
objective, the best anchored route is within 23--36\% of direct
optimisation, against 48--116\% without anchors---the measured cost of a
certificate that merges the extremes away. Missing evidence is not treated
as zero: a route with 82\% habitat coverage has its bound widened from 408
to 541 and wins no sampling.

No route is universally preferred: across 6{,}000 uniform weightings the
equal-weight May winner holds only a plurality, three of the four leaders
are anchors the certified search never returned, and the ice-free instance
concentrates (min~$d$ 77\%). Equity, read from geometry alone, disagrees
more with a set spanning more corridors: ties at small radii give way to
conflicts at $R{\geq}12$\,km in up to seven of fifteen cells, with the
social formulation undetermined wherever propagated ACS intervals overlap.
The approximation trade-off is sharp: $\varepsilon=0.25$ exceeds the
budget by 14.6$\times$, while $\varepsilon=0.5$ reduces the returned set
from 229 to five representatives. The approximate Pareto set alone is a poor decision set, up to 116\% off a
stated objective; the single-criterion anchors restore the corridors a planner
needs. Spatial deduplication keeps whichever route it encounters first and
once removed the ice optimum as a near-duplicate of a worse route, so anchors
are exempt. The deduplication threshold has no formal guarantee and changes
the winner across the tested settings. Route identifiers are positional, so
identity claims are paired with criterion values throughout.

The evaluation also highlights important limitations. Missing, measured-zero,
and non-discriminating evidence must remain distinct; available data may still
be uncertain or incomplete.
A limitation of the current system is that it does not support time-dependent multi-criteria routing.
\section{Social Impact}

The system makes community and
ecosystem exposure first-class inputs to a decision that today optimises
vessel risk and cost, on auditable open data, and lets a vessel's cargo and
size shape its route through standoffs a person chooses. Its value lies as much in restraint: declining to measure, to
rank and to merge are first-class outputs with their reasons.
A central ethical concern is that routing involves value judgments that should
not be made autonomously. Criterion weights and the definition of equity are
therefore supplied by user, while the system computes the resulting
trade-offs. Technical preference and equity are reported separately, so a
conflict between an operationally preferred route and a more equitable outcome
remains visible for human decision-making.

\begin{acks}
This research was funded by the National Science Foundation (NSF) under grant number 2531101. The authors also thank Andrew P. Barrett of the NSIDC, CIRES, University of Colorado Boulder, for his valuable suggestions and feedback as a member of the research team.
\end{acks}

\bibliographystyle{ACM-Reference-Format}
\bibliography{ref}

\end{document}